\documentclass[letterpaper, 10 pt, conference]{ieeeconf} 
\IEEEoverridecommandlockouts 
\usepackage{times}  % DO NOT CHANGE THIS
\usepackage{amsmath,amsfonts,mathtools}
\usepackage[hyphens]{url}  % DO NOT CHANGE THIS 
\usepackage{graphicx} % DO NOT CHANGE THIS
\usepackage{graphicx}  % DO NOT CHANGE THIS
\usepackage{amsmath, amssymb}

\usepackage{url}  
\usepackage{xfrac} 
\usepackage{color,soul} % remove later
\usepackage{gensymb}
\usepackage{multirow}
\usepackage{booktabs} % For formal tables
\usepackage{graphicx}
\usepackage{amssymb}
\usepackage{array}
\usepackage{amsmath}
\usepackage{lipsum}
\usepackage{amsmath}
\usepackage{paralist}
\usepackage[linesnumbered,ruled]{algorithm2e}
\title{\LARGE \bf SelfieDroneStick: A Natural Interface for Quadcopter Photography}

\author{Saif Alabachi$^{1}$ \quad Gita Sukthankar$^{2}$ \quad Rahul Sukthankar$^{3}$% 
\thanks{$^{1}$Saif Alabachi is with the  University of Technology, Baghdad, Iraq
        {\tt\small s.ghassan@gmail.com}}%
\thanks{$^{2}$Gita Sukthankar is with the Department of Computer Science, University of Central Florida, Orlando, FL
        {\tt\small gitars@eecs.ucf.edu}}%
\thanks{$^{3}$Rahul Sukthankar is with Google 
        {\tt\small sukthankar@google.com}}%        
}

\begin{document}
\maketitle
\thispagestyle{empty}
\pagestyle{empty}

\noindent
\begin{abstract}
A physical selfie stick extends the user's reach, enabling the acquisition of personal photos that include more of the background scene. Similarly, a quadcopter can capture photos from vantage points unattainable by the user; but teleoperating a quadcopter to good viewpoints is a difficult task.  This paper presents a natural interface for quadcopter photography, the \textit{SelfieDroneStick} that allows the user to guide the quadcopter to the optimal vantage point based on the phone's sensors.  Users specify the composition of their desired long-range selfies using their smartphone, and the quadcopter autonomously flies to a sequence of vantage points from where the desired shots can be taken. The robot controller is trained from a combination of real-world images and simulated flight data. This paper describes two key innovations required to deploy deep reinforcement learning models on a real robot: 1) an abstract state representation for transferring learning from simulation to the hardware platform, and 2) reward shaping and staging paradigms for training the controller. Both of these improvements were found to be essential in learning a robot controller from simulation that transfers successfully to the real robot.
\end{abstract}

\begin{figure*}[t]
\centering
	\includegraphics[width=0.9 \textwidth]{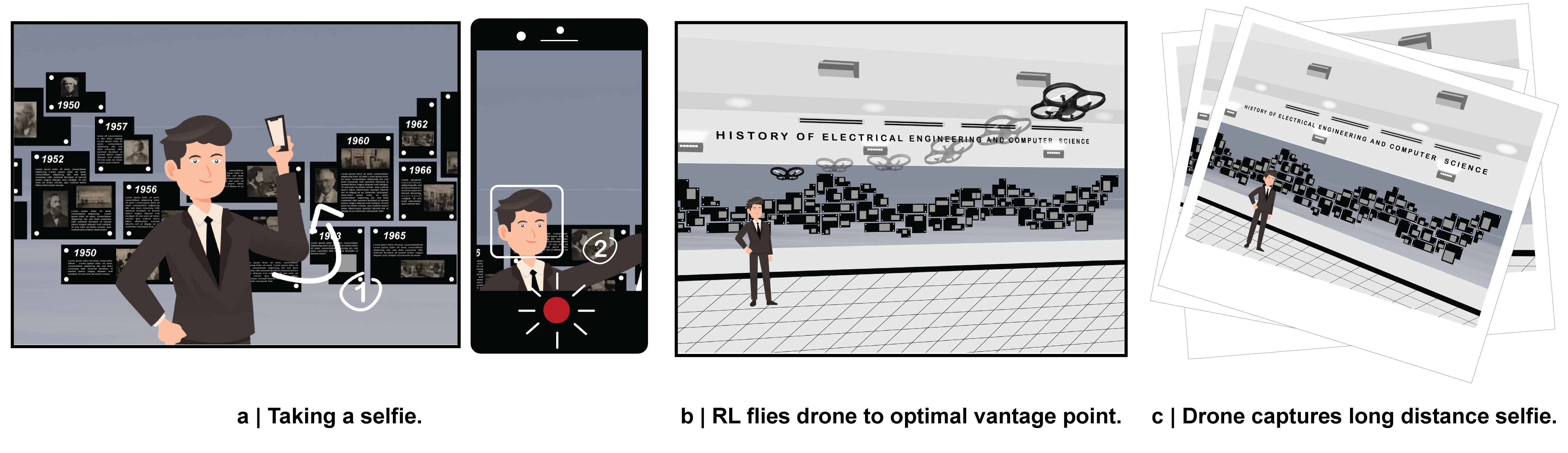}
\caption{\textit{SelfieDroneStick} enables long-range selfie photography by allowing the user to naturally specify vantage points from which a drone can capture well-framed photos of the user.
}
\label{fig:sds}
\end{figure*}

\section{Introduction}
% General Intro
Although there has been work on improving quadcopter teleoperation~\cite{lan2017xpose,alabachi2018intelligently} and robot-assisted photography~\cite{cheng2015aerial,Sukthankar-Rey-FLAIRS2016}, the premise behind most of these investigations has been that
the user must learn the proposed interface paradigm.  Our philosophy is to make the users learn as little as possible and the system learn as much as necessary.  The \textit{SelfieDroneStick} interface mimics the functionality of a selfie stick, enabling the user to control the quadcopter with only a mobile phone by a simple gesture and a click as shown in Fig.~\ref{fig:sds}. 

The goal is to generate a well-framed selfie of the user against the desired background, as if it were taken using a virtual selfie stick extending from the user in the direction of the handheld smart mobile device (SMD). The user specifies the desired composition by taking an ordinary selfie using the SMD, where the relative orientation directly specifies the azimuth of the vantage point while the height, 3D space position, and desired distance is indirectly specified by the SMD elevation, position and size of the user's face in the captured frame respectively. The drone flies to the target viewpoint based on the vantage point specified by the SMD to capture the selfie using a learned controller. The drone mirrors the bearing of the SMD as measured by its onboard IMU and selects an appropriate distance such that the user's body visually occupies relatively the same area in the drone selfie as the user's face did in the SMD frame. The resulting photos frame the user against the entire background, just as if the user had used a impossibly long selfie stick to compose the photograph.
 
%Perception is accomplished with DUNet, our new meta-architecture for real-time object detection.  The DUNet architecture consists of a sequence of dense blocks that process the input image at different scales, connected to a sequence of prediction layers, each of which independently generate detection results.  Two strengths of DUNet are its processing speed and ability to reliably detect small objects; in prior work, we demonstrate its ability to learn customized object detection models for mobile robots~\cite{alabachi2019customizing}. This learning strategy eliminates the necessity of having extra convolutional layers to account for variations in appearance and orientation.  
 
Instead of attempting to use deep reinforcement learning (RL) to learn a direct control policy based on the raw pixel data as was done in \cite{mnih2015human}, our controller utilizes an abstract state space representation. First, our perception system, Dense Upscaled Network (DUNet)~\cite{alabachi2019customizing}, is trained to detect a human face (which is prominent in the phone camera image) and also the human body (visible from the drone's viewpoint).
%We pre-train DUNet on PASCAL VOC~\cite{Everingham15} and WIDER FACE~\cite{yang2016wider} datasets for human and face object categories.
Deep Deterministic Policy Gradient (DDPG)~\cite{lillicrap2015continuous} is then used to learn the flight policy in simulation using an abstract, continuous state space before being transferred to the real robot.  To create a smooth and steady flight trajectory, we shape the reward to take into account both position and velocity.

This work introduces a novel interface for automating UAV selfie-photography using a mobile device. Our system takes selfies at the specified depth, background and orientation angle in the scene, with the user placed at the desired position in the frame.  Once trained, our RL controller autonomously flies the quadcopter to the user-selected vantage point. Creating a selfie using our system typically takes 5s for the user to compose the shot and 8s for the drone to fly the maneuver, vs.\ 60s to manually operate the drone for a similar shot. Our system architecture is shown in Figure~\ref{fig:sds}, and the ROS configuration, simulated environment, and code are publicly available.\footnote{%
%Download \textit{SelfieDroneStick} from
\url{https://github.com/cyberphantom/Selfie-Drone-Stick}}

\begin{figure*}[t]
\centering
	\includegraphics[width=0.7\textwidth]{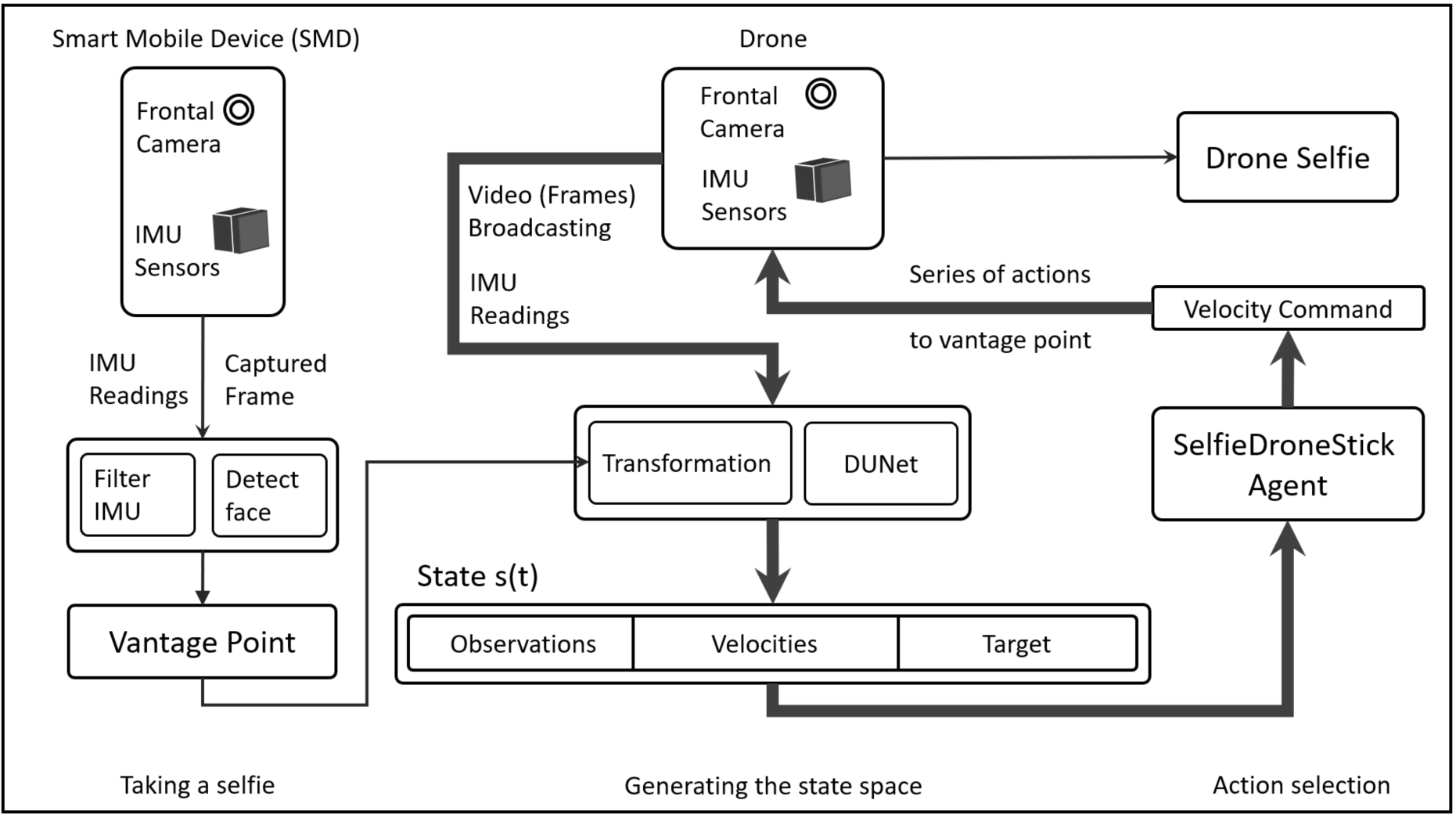}
\caption{Using the device camera, the user snaps the shot as if taking a selfie while angling the phone to modify the face size and background in the captured frame. The camera frame and IMU sensor readings are then extracted to generate the desired vantage point.  DUNet is employed to detect the human face and body in both the images captured by the mobile device and the drone.  The transformation module creates the target point, and the SelfieDroneStick software agent autonomously navigates there using the learned control policy.}
\label{fig:system_blockdiagram}
\end{figure*}

\section{Related Work}

% High-resolution cameras mounted on UAVs have motivated work on automating photography and videography~\cite{cheng2015aerial}, creating dynamic scenes from high and hard multi-view points~\cite{nageli2017real}, and recording cinematography videos following a virtual trajectory assigned through a 3D tool~\cite{joubert2015interactive}. However many tasks are too ill-specified to be performed autonomously, necessitating the development of specialized user interfaces.  

%Updated related work
%High-resolution cameras mounted on UAVs have motivated work on automating the performance of photography and videography similar to a professional  cameraman~\cite{cheng2015aerial, huang2019learning}, creating dynamic scenes from high and hard multi-view points~\cite{nageli2017real}, and recording cinematography videos following a virtual trajectory assigned through a 3D tool~\cite{joubert2015interactive}. However many tasks are too ill-specified to be performed autonomously, necessitating the development of specialized user interfaces. 

Natural user interfaces (NUI) rely on innate human actions such as gesture and voice commands for all human-robot interaction~\cite{popov2016control,fernandez2016natural, ma2016studies}.  Alternatively, more precise navigation in indoor and outdoor environments can be achieved through structured waypoint designation strategies~\cite{alabachi2018intelligently, gebhardt2016airways}.  Wearable sensors were employed in a
point-to-target interaction scenario to control and land a drone using arm position and absolute orientation based on the inertial measurement unit (IMU) readings~\cite{gromov2018video}.  Our system removes the need to employ gestures, hand crafted strokes, or wearable devices. Any mobile device equipped with a camera and IMU sensors can be used to direct the quadcopter using our \textit{SelfieDroneStick} interface.

A subset of the human-robot interaction research has specifically addressed the problem of user interfaces for drone-mounted cameras.   For instance, \cite{alabachi2018intelligently} tracks user-specified objects with an adaptive correlation filter in order to create photo collections that include a diversity of viewpoints.  XPose~\cite{lan2017xpose} is a touch-based system for photo taking in which the user concentrates on adjusting the desired photo rather than the quadcopter flight path. XPose offers an innovative, powerful user interface for taking a variety of photos, supported by trajectory planning.  For our research, we chose to focus on the most popular photo composition (selfie in front of sweeping background), while using deep learning to make the vantage selection process fast and intuitive for the user.

% Deep RL has been used to learn specialized flight controllers; for instance, Deep Q-Network (DQN) was used to learn autonomous landing policies for a quadcopter with a downward facing camera~\cite{polvara2017autonomous}.  It also has been employed for capturing frontal facial photos; in this work the authors~\cite{passalis2018deep} developed a realistic simulation for the task and trained their system on a database of head positions.   In order to use DQN, both these systems adopted a discretized action space.   In our system, we evaluated the performance of both DQN and DDPG and found that the DDPG produced trajectories with less oscillation, particularly when combined with our reward shaping method.

%Updated related work
Deep RL has been used to learn specialized flight controllers; for instance, Deep Q-Network (DQN) was used to learn autonomous landing policies for a quadcopter with a downward facing camera~\cite{polvara2017autonomous}. DQN also has been employed for capturing frontal facial photos; Passalis and Tefas~\cite{passalis2018deep} developed a realistic simulation for this task and trained their system on a database of head positions.  In order to use DQN, both these systems adopted a discretized action space.   In our system, we compare the performance of DQN, PG (policy gradient), DDPG (deep deterministic policy gradient) and found that the DDPG produced trajectories with less oscillation, particularly when combined with our reward shaping method. Rodriguez et al.~\cite{rodriguez2018deep} demonstrated good results with a similar approach on the challenging task on autonomous multirotor landing on a moving platform. In order to capture the perfect selfie, our learned controller must be able to execute multiple types of flight paths and is not limited to a single maneuver.

\section{Method}

Fig.~\ref{fig:system_blockdiagram} presents an overview of the \textit{SelfieDroneStick} system. First, the user specifies vantage points for the drone using an SMD, simply by clicking a series of selfies. For each vantage point, the system captures both a reference camera image as well as the SMD's corresponding orientation from its inertial measurement unit (IMU) sensor. By combining the orientation of the SMD with the framing of the user's face in the SMD camera image, we can extrapolate an ideal vantage point for the drone. This is transformed into a desired framing for the user in the drone's camera. Next, on the drone we combine information from its onboard IMU, its front-facing camera and the vantage point. We employ a fast object detector to localize the user in each frame (operating at frame rate is essential, particularly in a crowded scene) and form a state vector that is used by the SelfieDroneStick reinforcement learning (RL) agent to plan the drone trajectory to the next vantage point. Finally, the RL agent, which was trained in simulation, controls the drone via a series of velocity commands to guide it through the sequence of vantage points. When the drone reaches each vantage point, it captures a long-range selfie of the user. The following sections discuss each of these steps in greater detail.

\subsection{A Natural Interface for Vantage Point Specification}
\begin{figure}
\centering
\includegraphics[width=1.0\columnwidth]{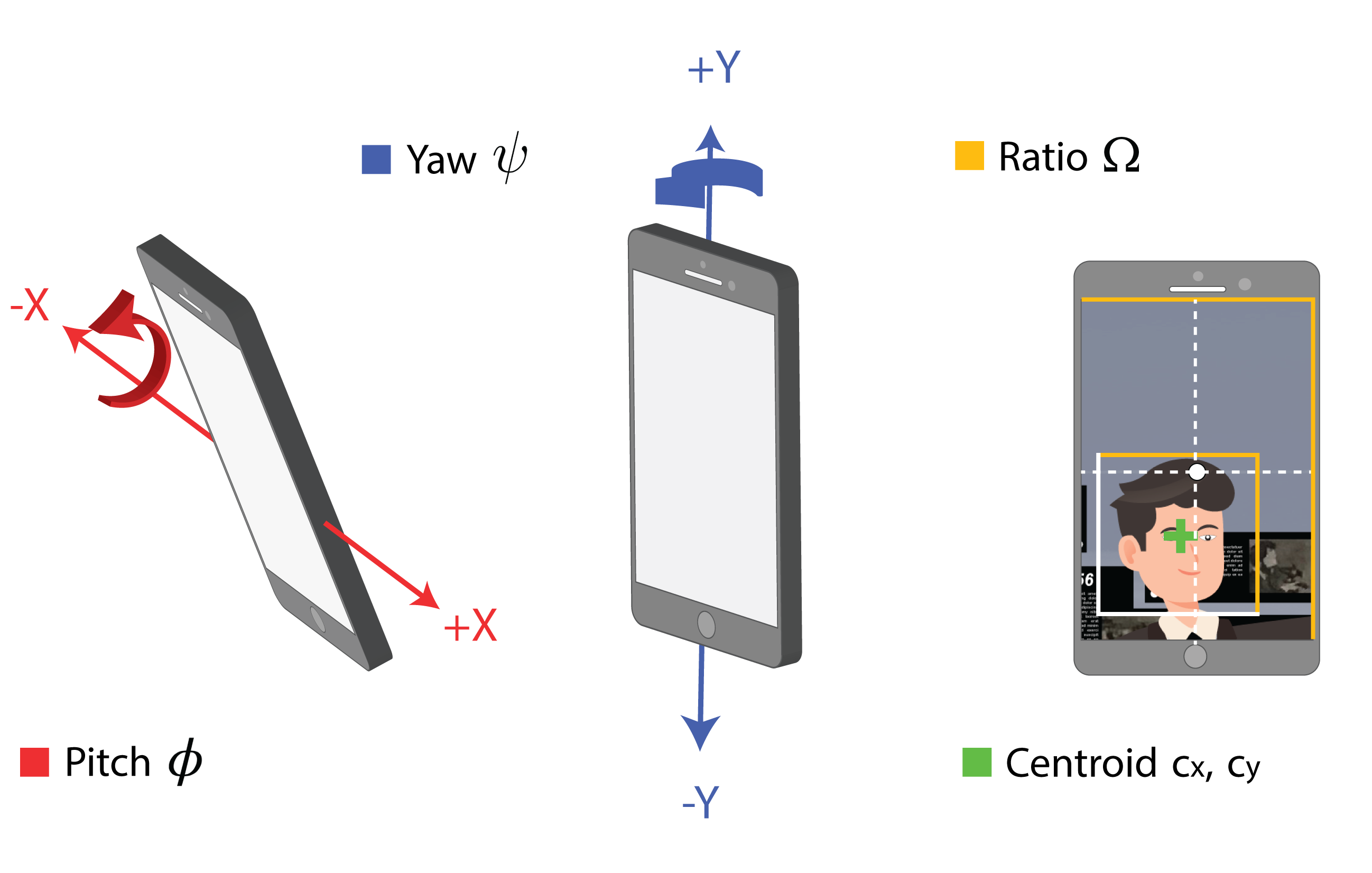}
\caption{The drone vantage point is specified by the orientation of the smart mobile device, $\phi$ and $\psi$, as well as the position and size of the user in the camera frame, ($c_x, c_y)$ and $\omega$. Moving the SMD away from the user corresponds to extending a ``virtual selfie stick'', guiding the drone to capture a photo from further away.}
\label{fig:trans}
\end{figure}

The user activates the SelfieDroneStick by taking a regular selfie using our web-based camera app. Then the shutter is pressed, the SMD's x-axis and y-axis orientation are recorded along with the camera image (Fig.~\ref{fig:trans}).
The IMU information partially specifies a bearing from the user along which the drone should seek to position itself in order to capture the desired shot. 
In addition to the bearing of the desired vantage point, we also need to specify the range. The key idea behind our interface is to enable the user to naturally specify the distance to the vantage point simply by varying the distance of the SMD from the user's face --- moving the SMD further away should cause the drone to capture photos from further away. Intuitively, extending the user's arm corresponds to a (magnified) extension of a selfie stick. This requires us to localize where the user is located within the selfie frame. We employ the DUNet architecture~\cite{alabachi2019customizing}, a real-time object detection CNN model, for face detection and the same model is also used for person detection on the drone camera (see below). The position of the user's face $(c_x, c_y)$, along with the ratio of the face bounding box to camera frame $\omega$ and the SMD orientation $\phi$ and $\psi$ fully specify the drone vantage point; this is illustrated in Fig.~\ref{fig:trans}.

How do these measurements translate to a specific drone position?  Visualizing the drone as if it were mounted on a virtual selfie stick extrapolated along the user's arm, past the SMD, shows that some of the coordinates map directly from the SMD to the drone: for instance, the azimuth to the drone is simply the yaw angle of the SMD, $\psi$. Others require an explicit transform: we empirically observe that the ratio of the user's face to the SMD image size falls in the range of $[0.1, 0.2]$ while desirable drone shots have a ratio of user's body to drone image size of $[0.03,0.4]$. Thus, we linearly map the ranges of the former to the latter to obtain suitable range distances for the drone. Finally, the height of the drone is derived from a combination of SMD tilt $\phi$ and position $(c_x, c_y)$ and size ratio $\omega$ of the user's face in the frame using straightforward geometry.

\subsection{Perception System}
There have been recent successes of Deep Reinforcement Learning that train end-to-end directly from pixel inputs in the context of automated game playing. However, for real-world robotics, it is still generally more practical to build reinforcement learning (RL) planners on top of stable perception systems.
We use a specialized object detector DUNet~\cite{alabachi2019customizing} that is trained for detecting small objects and optimized for inference on compute-constrained platforms.
%DUNet (Densely Upscaled Network) is a convolutional neural net (CNN) architecture that extends the meta-architecture of the popular SSD object detector with ideas from DenseNet, Feature Pyramid Network and Top-Down Modulation.

%At a high level, DUNet consists of a sequence of dense blocks that process the input image at different scales, connected to a sequence of prediction layers, each of which independently generate detection results. The two sequences are connected both laterally (at appropriate scales) and vertically (through max pooling and upscaling). To the best of our knowledge, DUNet is the first architecture to exploit both DenseNet-style concatenation (via dense blocks) in the bottom-up pathway and ResNet-style summation (via the upscaling) in the top-down pathway.

Since RL planners are trained in simulation, we employ an instance of DUNet pre-trained on data collected in the simulated world during training and replace it with a DUNet model pre-trained on PASCAL VOC~\cite{Everingham15} for real-world deployment. This enables the RL policy learned during simulation to transfer to the real-world without requiring explicit domain transfer. Since DUNet runs at real-time, we are able to perform per-frame detection while tracking the user through the stream of images. When the scene contains multiple people, it is important for the system to maintain its focus on the user; this is significantly easier in our application compared to the general tracking problem since the user is compliant and looking at the drone rather than averting their face or hiding behind obstacles. DUNet localizes detections with a bounding box and for the purposes of the RL planner, only the size and location of bounding boxes containing the user are relevant.

The drone's state space $s_t$ composed of: observations -- odometry derived from on-board sensors (gyroscope, accelerometer and pressure readings) and localization of the user in the drone's front-facing camera (generated by DUNet), (b) linear and angular velocities, (c) target -- vantage point specification.

When designing a state space, it is worthwhile to focus only on the relevant features since RL scales poorly with state space dimensionality. For this reason, we condense (a) to a 5-D tuple that specifies the drone's pose relative to the user: $\mathbf{a}=[\psi^d, \Upsilon^d, c_x^d, c_y^d, \omega^d]$, corresponding to the current azimuth to the user, drone height, and observed bounding box location and size ratio, respectively; the superscript $^d$ denotes that these are all measurements on the drone rather than similar parameters measured on the SMD.

The velocity state (b) is straightforward, $\mathbf{b}=[\dot{x},\dot{y},\dot{z},\dot{Z}$
consisting of three linear velocity components and the angular velocity around the vertical axis (drone yaw rate of change).

The final aspect of the state (c) is the location of the next vantage point, specified using the same coordinates as the drone pose: $\mathbf{c}=[\psi^v, \Upsilon^v, c_x^v, c_y^v, \omega^v]$, as above with superscript $^v$ denoting that this specifies the vantage point.

The complete state vector is a concatenation of these three tuples, $\mathbf{s}_t = [\mathbf{a}, \mathbf{b}, \mathbf{c}]$, resulting in a 14-dimensional state space.

\subsection{Learning a Deep RL Robot Control Policy: Rewards}

At a high level, the goal of the \textit{SelfieStickDrone} agent is to pilot the drone quickly and smoothly to each vantage point, without overshooting or oscillating. A naive formulation of such a goal in reinforcement learning (RL) would be to place sparse rewards only at the vantage points. Such an approach can work for simple problems in simulated environments but is challenging on our task because RL systems are inherently high in sample complexity and this is exacerbated by environments with sparse rewards (confirmed in our experiments below).

Practical RL for robotics relies heavily on training RL policies in simulation and then transferring the learned models to the real world. It also benefits significantly from \emph{reward shaping} and curriculum learning through \emph{staging} of rewards. We describe these in the context of our application.

Reward shaping for RL requires a careful balance between terms that are so punishing that they drive agents to absorption states (e.g., penalizing the agent at each time step in order to incentivize efficient flights could encourage the agent to end its episode quickly by crashing) and reward functions that are too rewarding near the goal that the agent chooses to dawdle near the goal state, accumulating a long sequence of partial rewards without achieving its objective.

Our reward function consists of two main terms: (1) a basin of attraction surrounding the specified vantage point to incentivize policies that fly the drone to the goal, and (2) a term to punish high-speed flight near the vantage point to encourage a smooth, non-oscillating approach before taking the selfie. The reward function also considers the fact that the drone flies through a sequence of vantage points and must smoothly transition from one to the next.

While creating a good reward function required considerable experimentation, we can explain its construction intuitively as follows. A natural expression for distance from the current drone pose to the next vantage point (goal) is given by the Euclidean distance between the corresponding pose vectors: $||\mathbf{a}-\mathbf{c}||_2 =
 || (\psi^v, \Upsilon^v, c_x^v, c_y^v, \omega^v) -
    (\psi^d, \Upsilon^d, c_x^d, c_y^d, \omega^d) ||_2$.
We want the reward to decay with distance from the goal, and we also want to penalize speed when near the goal. From these, we propose the following formulation for the reward:
\begin{equation}
R = \text{CLIP}_{[0,1]} \left( \cos(\gamma ||\mathbf{a}-\mathbf{c}||_2 e^{-\alpha ||\mathbf{a}-\mathbf{c}||_2}) e^{-\beta ||\mathbf{b}||_1}
\right) ,
\end{equation}
where $\text{CLIP}_{[0,1]}(.) = \max(\min(., 1), 0)$, $\{\mathbf{a}, \mathbf{b}, \mathbf{c}\}$ are the three components of the RL state vector (described above) and $\{\alpha=1.3, \beta=0.35, \gamma=11.3\}$ are empirically tuned hyperparameters. The $\cos(.)$ term bounds the basin around the vantage point where the agent can collect partial rewards and high speeds are not rewarded near the goal.

Each episode runs until one of the following termination conditions is triggered:
\begin{itemize}
    \item If the agent achieves a reward of $>0.85$ in a given timestep, the episode is terminated as an early success, with a reward $+1$.
    \item If the drone flies outside the safe zone (exceeds height or ratio limits), the episode is terminated as an early failure, with reward $-0.8$.
    \item If the object detector fails to find the subject (e.g., user is out of view), the episode is terminated early, with reward $-0.8$.
    \item If the step counter reaches the max.\ episode length (41 in our experiments), the episode terminates with the current reward, $R$.
\end{itemize}
At each time-step, the agent receives a reward $R$ except for two cases, where the reward is explicitly shaped:
\begin{enumerate}
    \item An exploration reward of $+1$ is given whenever the agent achieves $R>0.75$; this is to encourage it to explore nearby states to achieve early success.
    \item When the drone moves so that the detected person falls very close to the edge of the image (within 10 pixels of frame), the reward is set to $-0.8$, but the episode is not terminated early; unless the drone acts quickly, the person will go outside the frame and trigger the early condition discussed above.
\end{enumerate}
The reward shaping incentivizes the agent to keep the user in its field of view and to stay within the safe zone. Once it can achieve these basic objectives, the agent can learn to fly the drone towards the vantage point.

A few other subtleties are worth mentioning: we provide the maximum reward $+1$ whenever the drone is within a radius of $1m$ of the vantage point and within $\pm 10^\circ$ in orientation. This acknowledges that the user's specification of the vantage point is intrinsically approximate and encourages the drone to fly quickly to the vantage point and take selfies while hovering rather than making unnecessary minor adjustments in an attempt to hit the exact position, which would have no meaningful impact on the qualities of photos acquired. The zero reward that the agent receives when it is far away from vantage points encourages the drone to fly quickly through those regions and the penalty for failing to detect the user is an incentive to keep the user in view (when possible). The latter is a form of curriculum learning: the drone first learns to explore while keeping the user in view and then learns to head to the vantage point -- without losing the user.

%Positive reward is used to encourage the agent to keep going when it is in the most meaningful region in the state space. $r(t) = +1.0$ reward is given when the error rate $er\le 0.15\%$ which is about a radius of $1.0 m$ from the exact vantage point position and $\pm10^\circ$ error in orientation. However, the agent gets penalized with $r(t) = -1.0$ when it is end up the episode with a failure terminal conditions such as missing the human object in successive steps more than the assigned threshold or when the drone height or the human body ratio  exceeding the the assigned threshold limitations. Otherwise, the agent is getting zero reward when it is far away from the target to encourage the agent to reach to the target as quickly as it can and avoid the conflict between the adjacent targets. With this procedure, the agent learns to stick with the detected human body first before creating a smooth gradient for reaching the optimal policy.

%\begin{figure}
%\centering
%\includegraphics[width=1.0\columnwidth]{rwd_explained.png}
%\caption{XXX- This figure will help to understand Fig.~\ref{fig:rwd} }
%\label{fig:rwd_explained}
%\end{figure}

%\begin{figure}
%\centering
%\includegraphics[width=1.0\columnwidth]{rwd_shaping.png}
%\caption{Shaping the reward function for distance vs.\ velocity to achieve a stable trajectory and a good quality long-range selfie by the drone.}
%\label{fig:rwd}
%\end{figure}

\begin{figure*}[t]
\centering
	\includegraphics[width=0.8\textwidth]{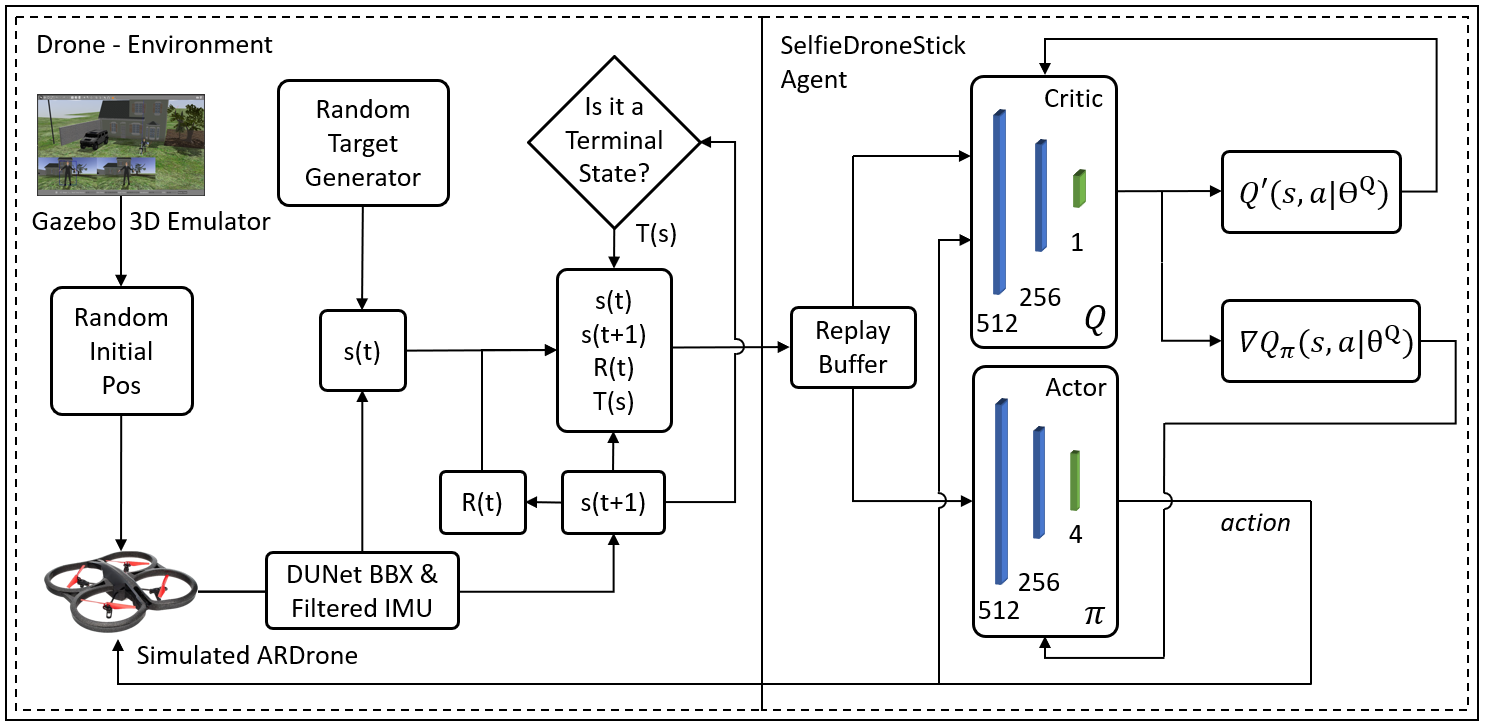}
\caption{Overview of the Deep RL agent training pipeline (target networks not shown). During each episode, the simulated drone is initialized in a random pose and assigned a random vantage point as target.
%The training phase of our SelfieDroneStick agent. In each episode, the simulated drone starts from random positions and yaw angle inside the environment to increase exploration margins. Random targets generator is used to set a new target for each episode.
The target critic network predicts the Q-value ($Q'$) and the critic network provides the action gradients ($\nabla Q_{\pi}$).
}
\label{fig:selfieDroneStick_training}
\end{figure*}

%\subsection{SelfieDroneStick Agent:}
\subsection{Deep RL: Design Choices and Real-World Transfer}
% we can also name it:\subsection{Learning Optimal policy}

\begin{figure}
\centering
	\includegraphics[width=0.49\textwidth]{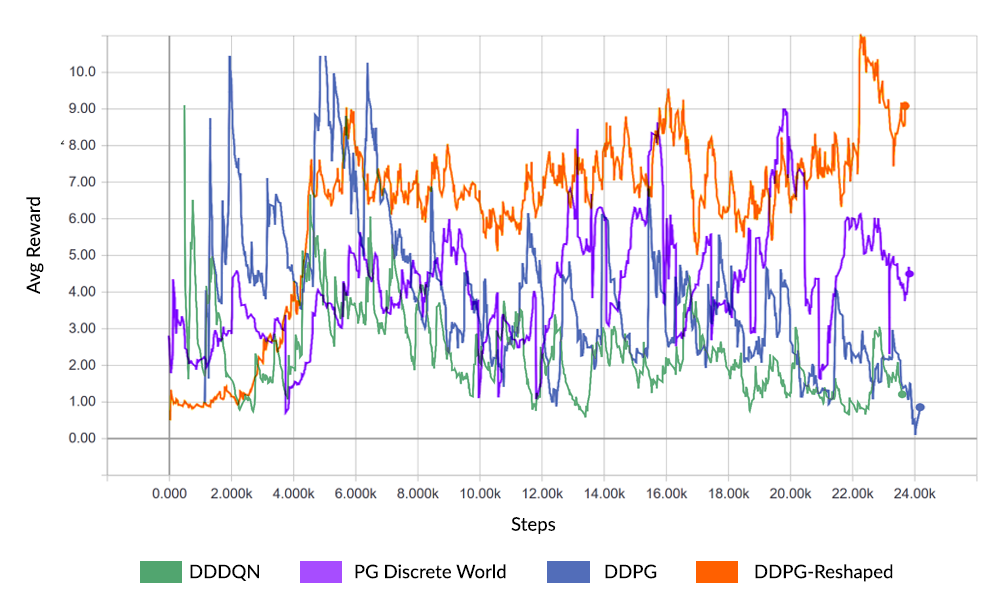}
\caption{
Training progress for different Deep RL methods. DDPG with reward shaping improves steadily.
%\textcolor{red}{training different state-of-the-art algorithms with the same environmental configuration. Deep Deterministic policy Gradient (DDPG) model shows a progressive and stable training after reshaping the reward function.}
}
\label{fig:methods}
\end{figure}

It is challenging to train a deep reinforcement agent to fly a real-world drone from scratch on this task. Therefore, we trained our system in a 3D emulator with a custom-built environment and a physical simulation of our quadcopter. The emulator creates a series of training simulations for the SelfieDroneStick agent with different initialization poses and vantage points. The agent tries actions, observes a new state and collects rewards at each episodic time-step $t$. We explored training the agent with a variety of Deep RL algorithms (described below) but at a high level, the goal is to learn a policy $\pi(\mathbf{s})$ by updating a value function $Q_\pi(\mathbf{s}, \mathbf{a})$, where $\mathbf{s}$ and $\mathbf{a}$ denote the agent's current state and its selected action, respectively. Fig.~\ref{fig:methods} illustrates the learning progress for each of the tested methods. 

%Reaching the vantage point in the 3D world requires a trained agent that is able to predict a chain of command velocities. However, it is almost impossible to train an agent to achieve this task in real-world with a physical quadcopter. Therefor, a 3D emulator with a simulated-version of the used quadcopter and a custom-built environment is used for training to find the optimal policy $\pi(s)$ through updating its value function $Q_\pi(s,a)$ by receiving higher rewards$r(t)$. Our agent trains itself, tests new action, observes new state, and collects rewards at each episodic time-step $t$.

% In spite the fact that deiscretizing the action space is the most common procedure that is followed by researchers to train with recent RL algorithms such as deep Q network and policy gradient method. However, this leads to having an infinite number of actions if we consider multiple velocities which in result discourage our NN convergence if it is done finely, and if considering wider margins in an attempt to reduce the action space size, the agent movement loses stability and we will have oscillated movement in the episode trajectory. 

Our action space is 4D and continuous: 3 linear velocities ($\dot{x}, \dot{y}, \dot{z}$) and a yaw rate $\dot{Z}$. Typical RL agents in the literature select an action from a discrete set (e.g., video game playing RL agents that select which button to press on a game controller) and learning in a multi-dimensional continuous action space, as is common in robotics, is more challenging.

Fig.~\ref{fig:selfieDroneStick_training} illustrates the training phase.  We consider and evaluate several recent approaches for model-free Deep RL that have achieved state-of-the-art performance across relevant domains, as detailed below.
%: DDDQN, PG, DDPG and DDPG with reward shaping, as detailed below.

% We focus on a learning a model-free off policy to be able to transfer the learned experience among different environments. Learning a neural network from a wide state space such as 3D space for the optimal policy is hard due to the high variance especially if we want to consider dynamic linear and angular velocities which is important for reaching the vantage point as fast as possible without loosing our reference. We investigate the most recent state-of-the-art Reinforcement Learning algorithms and we are able to achieve convergence by supporting our learning procedure with our proposed reward shaping algorithm.

\subsubsection{Dueling Double Deep Q Network (DDDQN)}

DDDQN is an updated variant on the popular Deep Q Network (DQN)~\cite{mnih2013playing}. Like DQN, this algorithm requires a discretization of the action space, for which we employ a simple 3x3x3 grid. The `double' refers to the use of two independent networks to address the over-estimation of $Q$ values in original DQN~\cite{van2016deep} and the `dueling' splits the network into two separate parts~\cite{wang2015dueling}: one for estimating the advantage of selecting a given action among the others for the given state, and a second for estimating the state value.

%\textcolor{red}{
%Dueling~\cite{wang2015dueling}
%double DQN~\cite{van2016deep}
%First, we use Dueling Double Deep Q Network (DDDQN) to learn from a discrete action space for each state in the proposed trajectory. This engages two independent networks to handle Q-values overestimation problem. In Dueling ,the network is separated into two networks for estimating the advantage of selecting each action among the others for the state and the other network is used to estimate the state value to boost the performance. However, we are still limited to use discrete actions space which limits the speed of the drone and if wider action space is considered, we will not be able to reach the optimal policy as shown in Fig.~\ref{fig:methods}. 
%}

\subsubsection{Policy Gradient} \label{sec:ddpg}

Policy Gradient (PG) is an on policy approach for learning stochastic policies.  In order to find the optimal policy, it increases the probabilities of actions that lead to higher return and reduces the probabilities of actions that lead to a lower return. During our evaluation of stochastic Gaussian policy gradient, we experienced convergence problems, accompanied by noisy gradients and high variance. Thus, we also considered the popular augmentation, Deep Deterministic Policy Gradient (DDPG)~\cite{lillicrap2015continuous}, which is based on the prior work on DPG~\cite{silver2014deterministic}.

\subsubsection{Deep Deterministic Policy Gradient (DDPG)}
 
DDPG is an off-policy, actor-critic algorithm that yields good environment exploration through the use of a stochastic behavioral policy.  The intuition is that it is easier to learn the optimal $Q$ value by employing greedy deterministic policy learning through following the time difference (TD) bootstrapping error. The actor takes the state and predicts the action using its policy network, and the critic provides the $Q$ value (expected return) based on the state and the actor's predicted action. Optimizing the $Q$ value of the critic network is done by minimizing the loss between the prediction of the critic target network and the expected return.
We implemented DDPG with a continuous action space that enables the drone to fly at a velocity of up to $4m/s$ in each direction.

\subsubsection{DDPG with shaped reward}
As above, but with reward shaping to encourage faster training. In the earlier variants, the reward function is ablated to consider only the sparse terms for vantage point proximity.

\section{Implementation Details and Experiments}

Transferring the learned controller from simulation to the real world is simplified by our perception system and design of the state space. The Gazebo simulator and ROS were configured to work with the OpenAI Gym toolkit to observe a new state every 160ms. Training was performed on a single NVidia Titan~X GPU.  To expedite the training, we set threshold values for the human object ratio and drone height to generate a safe zone for the drone movement. The vertical space is set to be in the range of $[0.5, 3]$ meters and the yaw angle to be between $[-75^\circ,75^\circ]$. To prevent the quadcopter from being too close to the detected human or moving too far and losing references, the human size ratio is only valid in the range $[0.5,0.03]$.
%The 10 pixels from each side of the quadcopter frame which are $[0-10, 630-640]$ horizontally and $[0-10, 350-360]$ vertically are set as thresholds to avoid losing the human subject during trajectory. If the drone falls into any of these thresholds during the training steps, we penalize the agent for that step $R(t)=-1$ and terminate the episode. 

%Network hyper parameters and structure are set to meet the wide range of states in the continuous state space and to compensate for the multiple target goal assignments. 
Both the actor and critic networks have two hidden layers with sizes $[512, 256]$. The Adam optimizer is used for training with the learning rates set to $10^{-5}$ and $10^{-3}$ respectively.
%Elu activation function and the batch-normalization layers are eliminated as their effect is very minimal with such low capacity network. 
The output action selected by the actor is scaled to be in the range of $[-0.8,0.8]$ as we are using a commodity quadcopter that exhibits shaky motion when flying faster than 0.8 ($\simeq 4m/s$) in any direction; this camera shake results in significantly degraded image quality and detection performance. The Deep RL agent is trained for 18K epochs in our simulated environment and then deployed on the drone.

%with 30fps camera frame-rate where a shaky movement in a speed above 0.8 $\simeq 4 m/s$ in any direction distort the the appearance of the human object in the frames resulting in degraded detection performance. With these settings, \textit{SelfieDroneStick} agent is trained after 18K epochs and Fig.~\ref{fig:tensorboard} illustrates the training progress and our \textit{selfieDroneStick} agent is able to generate trajectories for flying through a sequence of multiple vantage points in a single forward density model. 

%We assigned multiple targets in the simulated world and test the agent performance when it is trained with a linear reward function based on the distance to vantage point vs.\ our shaped reward function.  Table~\ref{tab:stats} summarizes the improvement from reward shaping

%performance of our approach and Fig~\ref{fig:paths} illustrates the trajectory in achieving the vantage point.
% sixth vantage point of Fig.~\ref{fig:targets}. 

\begin{figure*}[t]
\centering
	\includegraphics[width=0.7\textwidth]{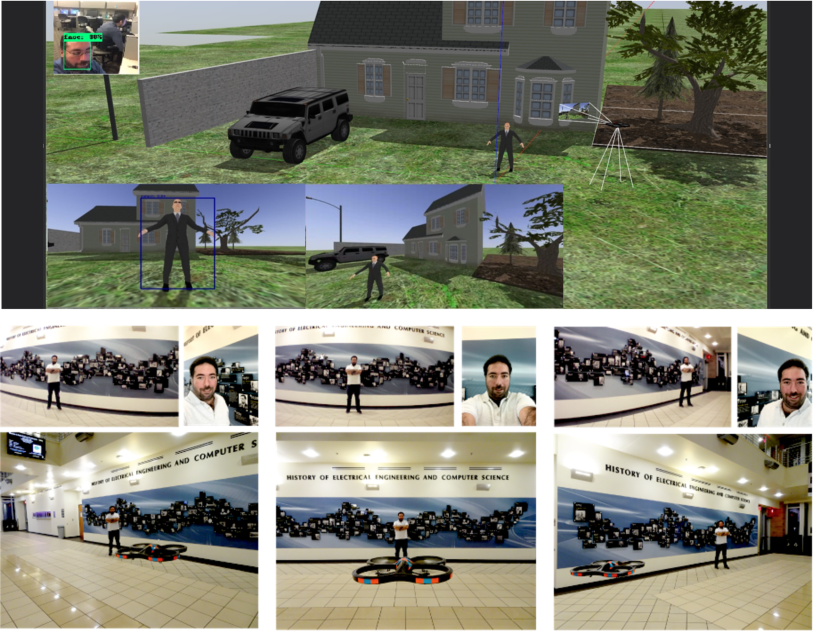}
\caption{Evaluating \textit{SelfieDroneStick} in a custom simulation environment and in real-world using the Parrot ARDrone 2.0. }
\label{fig:real}
\end{figure*}

During training, our environment consists of a 3D domain containing a single human and single UAV (drone) model against a simple background. We test the system in several realistic simulated scenarios, such as the one shown in upper part of  Fig.~\ref{fig:real}. This allows us to conduct end-to-end experiments under repeatable conditions with known ground-truth, with the same perception system as we employ for real-world experiments, as shown in the lower part of Fig.~\ref{fig:real}.

Both the simulated and real-world scenarios follow a consistent script:

\begin{enumerate}

\item The drone is initialized facing the human subject at a safe distance. In simulation, take-off and landing are straightforward; for the real drone, the user initiates take-off by holding the SMD flat and initiates a landing sequence by tilting the SMD $90\degree$ around the $x$-axis.

%Take-off is autonomously in the training phase with random target and initialization point, whereas it is activated by holding the SMD with $\simeq 0\degree$ degree alignment around $(x, y, z)$ coordinates, and the landing command is sent by tilting it $\simeq 90\degree$ around the $x$-axis. 

\item Prior to activating the SelfieDroneStick, the drone hovers in front of the user, centering the user in the middle of the image with $\Omega_{obs}\simeq24\%$.

%\item In training, the assigned target point is generated randomly and there is no translation process before our agent activation to speed up the learning. In testing, once the user takes a selfie with the SMD, the assigned vantage point is translated for the long-range selfie target..

\item Once the SelfieDroneStick agent has been activated by the user taking a selfie using the SMD, the SelfieDroneStick agent flies the drone to the specified vantage point using the learned Deep RL controller.

\item Once the drone arrives at the bearing and range consistent with the specified vantage point, it takes a long-range selfie of the subject.

\end{enumerate}

%\begin{figure}[t]
%\centering
%	\includegraphics[width=0.47\textwidth]{rwd_stat.png}
%\caption{Episodic rewards, drone shots counter,  Q-value, and \# steps per episode are monitored and recorded during our training. Our reward function incentivizes the RL agent to first focus on the user. Then it learns to fly the drone to the specified vantage point from a variety of initializations. Training converges in 19K epochs, after which we transfer the agent from simulation to the physical drone.

%First, \textit{SelfieDroneStick} agent learns to stick with the target object. Then, it learns to fly the drone to the assigned vantage point with a random initialization starting point. After only 19K epochs, our agent achieves the highest reward which indicates that it is learned and ready to be tested.
%}

%\label{fig:tensorboard}
%\end{figure}

% No reference. Figure removed
%Fig.~\ref{fig:tensorboard} shows the obtained reward after learning from the simulated data. The learned controller is then evaluated in realistic simulated scenarios (Fig.~\ref{fig:real}).

% \begin{figure}[t]
% \centering
% 	\includegraphics[width=0.47\textwidth]{targets}
%     %\includegraphics[width=0.4\textwidth]{fig6}
% \caption{
% Samples drawn from the simulated training environment in which the Deep RL agent is trained to fly against varying backgrounds from different vantage points.
% }
% \label{fig:targets}
% \end{figure}

\begin{figure}
\centering
	\includegraphics[width=\columnwidth]{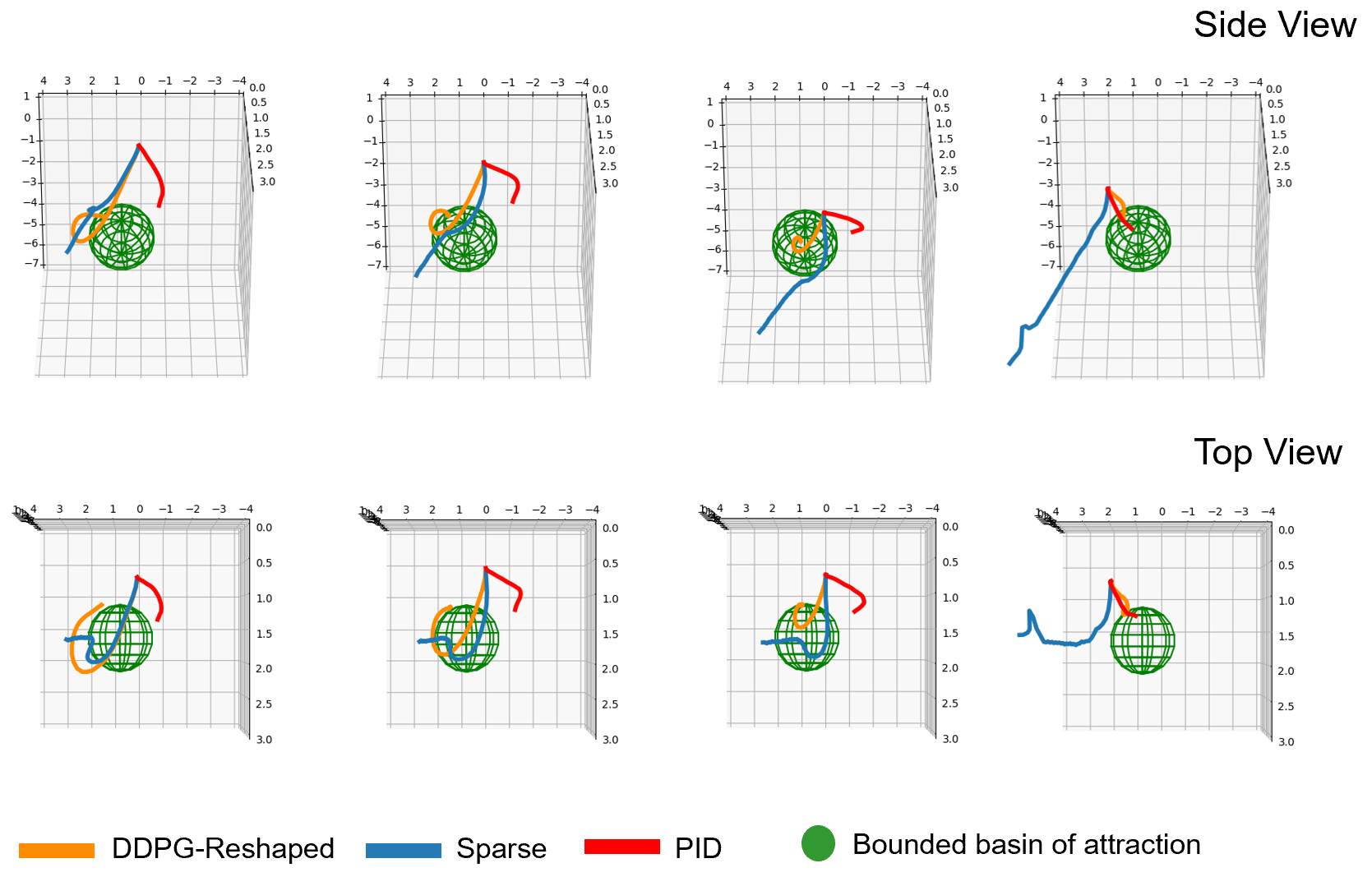}
	\caption{Proposed controller (DDPG-reshaped) compared against a PID and DDPG trained on sparse rewards on 4 drone photography scenarios. The green sphere denotes target vantage point. See text for details.}
    %\includegraphics[width=0.4\textwidth]{fig6}
    % PID added here. I think we need to change the caption here.
%\caption{Importance of reward shaping: with sparse rewards, there is low consideration of the speed when reaching the vantage point while the shaped reward guides the drone to the vantage point and reduces the speed ahead of a time until hovering state. PID scenario is also tested as an additional unlearning control baseline. In PID, we experience high probability for the drone in loosing its reference.}
%The drone trajectory to reach the vantage point with standard reward function vs. our reshaped reward function.
\label{fig:paths}
\end{figure}

Fig.~\ref{fig:paths} illustrates drone trajectories generated by the proposed controller (DDPG-reshaped) against two baselines, a classical PID contoller and a DDPG controller trained on sparse rewards. We see top and side views (rows) for each of four scenarios (columns) as the drone approaches the vantage point (green sphere). The shaped reward is better at slowing the drone to a stable hover at the vantage point. DDPG with sparse rewards fails on the fourth scenario.

To better understand the effects of reward shaping, we compared the proposed shaped reward against a baseline reward that varies linearly with distance to vantage point. We measure both the distance to target and the variance in velocity over 10 timesteps to assess the controller's ability to bring the drone to the vantage point and its stability, respectively (see Table~\ref{tab:stats}).
\begin{table}
\caption{Evaluation of proposed reward shaping vs.\ baseline}
\begin{center}
%\begin{tabular}{p{0.5cm}|p{0.5cm}p{0.5cm}|p{0.5cm}p{0.5cm}}
\begin{tabular}{c|cc|cc}
%     \multirow{2}{2pt}{Init.\ Start Point} & \multicolumn{2}{c}{Std.\ Rwd f.} & \multicolumn{2}{c}{Reshaped Rwd f.} \\
Scenario & \multicolumn{2}{c}{Baseline Reward} & \multicolumn{2}{c}{Shaped Reward}\\
    & Dist & {Var. Vel} &
    Dist & {Var. Vel} \\
    \midrule
    1 & 0.133 & 0.168 & 0.144 & \textbf{0.141} \\
    2 & 0.235 & 0.0707 & \textbf{0.148} & 0.148 \\
    3 & 0.089 & 0.155 & 0.100 & \textbf{0.147} \\
    4 & 0.124 & 0.170 & \textbf{0.085} & \textbf{0.144} \\
    \hline
\end{tabular}
%\caption{The mean SSE is for 41 steps per episode. The variance are calculated for the last 10 steps to measure the stability of the drone when reaching the vantage point (Each episode is assigned to have 41 steps).}
\end{center}
\label{tab:stats}
\end{table}

Finally, Fig.~\ref{fig:real} shows the \textit{SelfieDroneStick} interface operating in three real-world scenarios. These experiments employed an iPhone SMD in conjunction with an AR.DRONE 2.0 UAV with a 30fps frame rate, with video streamed to a laptop that runs percepion and control. The interface enables the user to take multiple selfies with different backgrounds as the user moves in the environment.  Videos of the system can be viewed at \url{http://ial.eecs.ucf.edu/SelfieDroneStick/}. 

% From the results, we can claim that using the baseline controller works efficiently if we want to reach the target faster but that may lead to losing the detected human body in the middle of the trajectory and hence the drone losses its reference. However, the trained SDS agent has already seen such cases and got punishment reward, so in the testing face we aren't subjected to losing the human body anymore.

\section{Conclusion}
This paper introduces the SelfieDroneStick, our autonomous navigation and selfie-photography platform that takes long-range selfies using a drone from vantage points specified by the user using a natural ``virtual selfie stick'' interface.
%This paper describes the most significant lessons we learned from designing the system. 
Designing the SelfieDroneStick interface required overcoming several significant challenges: (1) specifying the composition of desired selfies using the smartphone, (2) learning Deep RL policies that transfer from simulation to the real world robustly, (3) ensuring that perception, cognition and control operate on compute-constrained platforms at frame-rate. Our experiments in simulation and on the quadcopter confirm the feasibility of creating a natural interface for quadcopter photography driven by a learned RL policy.

%Deep learning can help make the user experience more natural; rather than having the photographer learn how to use the system, the system should learn the flight controller that best expresses the user's wishes. Although our current research utilizes reward shaping to express human preferences, in future work we are planning to employ learning from demonstration as an alternative to reward shaping.

%from the desired yaw angle, height and depth, and the spatial position of the foreground subject with respect to the background. Three cooperative agents handle SDS tasks sequentially starting from target assignment and ending with learning a DNN for action prediction. SDS uses any SMD supported with a camera and IMU sensors to create the target photography viewpoint and predict the correct sequence of actions that achieve it. Our platform can navigate to multiple targets in a single low-capacity DNN through the relevant and generic state space. Our system has been tested with different simulated models and in real world.

% Drawback
%The draw back for this implementation coming from changing opposite actions that makes the UAV oscillates and gambles which slows down the learning due to losing the lane of sight to the subject. We compensate for this draw back by reducing the speed of the UAV. However, it is more likely to solve this problem by using advanced drone with gambling free camera and higher frame rate.

\section*{Acknowledgments}
%The authors would like to thank Yasmeen Alhamdan for assisting with figure generation.
\noindent
We thank Yasmeen Alhamdan for help in figure generation.

\bibliography{bib}

\end{document}